\documentclass[10pt,twocolumn,letterpaper]{article}

\usepackage{iccv}
\usepackage{times}
\usepackage{epsfig}
\usepackage{graphicx}
\usepackage{amsmath}
\usepackage{amssymb}
\usepackage{booktabs, makecell, tabularx}
\usepackage{multirow}
\usepackage{dsfont}
\usepackage{enumitem}
\usepackage{algorithm}
\usepackage{algpseudocode}
\usepackage{xspace}

\usepackage{animate}
\usepackage{booktabs} 

\usepackage{booktabs}
\usepackage[normalem]{ulem}
\usepackage{ifthen}
\usepackage{epsfig}
\usepackage{graphics}
\usepackage{graphicx}
\usepackage{amsmath}
\usepackage{animate}
\usepackage{amsmath}
\usepackage{booktabs}
\usepackage{hhline}
\usepackage{adjustbox}
\usepackage{multirow}
\usepackage{booktabs,arydshln}
\usepackage{amsfonts}

\usepackage[pagebackref=true,breaklinks=true,colorlinks,bookmarks=false]{hyperref}

\iccvfinalcopy 

\usepackage{color}
\usepackage[dvipsnames]{xcolor}
\usepackage{caption}

\newcommand{\reffig}[1]{Figure~\ref{fig:#1}}
\newcommand{\refsec}[1]{Section~\ref{sec:#1}}

\newcommand{\reftbl}[1]{Table~\ref{tab:#1}}

\newcommand{\lblfig}[1]{\label{fig:#1}}

\definecolor{MyDarkBlue}{rgb}{0,0.08,1}
\newcommand{\camera}[1]{#1}

\newcommand{\myparagraph}[1]{\paragraph{#1}}
\newcommand{\feature}{\bs{f}_{t}\xspace}

\newcommand{\website}{\href{https://text2cinemagraph.github.io/website/}{\textit{website}}\xspace}

\newcommand{\bs}[1]{{\boldsymbol{#1}}}
\newcommand{\pixel}{p\xspace} 
\newcommand{\art}{\bs{x}\xspace} 
\newcommand{\nat}{\hat{\bs{x}}\xspace} 
\newcommand{\artt}{\bs{x}_t\xspace} 
\newcommand{\natt}{\hat{\bs{x}}_t\xspace}  
\newcommand{\arttminus}{\bs{x}_{t-1}\xspace}  
\newcommand{\nattminus}{\hat{\bs{x}}_{t-1}\xspace}  
\newcommand{\atten}{\bs{A}_t\xspace}  
\newcommand{\attenavg}{\bs{\overline{A}}\xspace}  
\newcommand{\artmask}{\bs{M}\xspace}   
\newcommand{\natmask}{\hat{\bs{M}}\xspace}  
\newcommand{\artflow}{\bs{F}\xspace}  
\newcommand{\natflow}{\hat{\bs{F}}\xspace} 

\newcommand{\at}{\bs{c}\xspace}   
\newcommand{\nt}{\hat{\bs{c}}\xspace} 
\newcommand{\gf}{\bs{G}_{flow}\xspace}
\newcommand{\gv}{\bs{G}_{frame}\xspace}

\definecolor{myblue}{rgb}{0.239,0.553,0.565}

\definecolor{cmu_red}{rgb}{0.706,0.169,0.212}

\begin{document}

\title{Text-Guided Synthesis of Eulerian Cinemagraphs}


\author{Aniruddha Mahapatra\textsuperscript{1}
\qquad
Aliaksandr Siarohin\textsuperscript{2}
\qquad
Hsin-Ying Lee\textsuperscript{2}
\qquad \\
Sergey Tulyakov\textsuperscript{2}
\qquad
Jun-Yan Zhu\textsuperscript{1}\\
\\
\textsuperscript{1}Carnegie Mellon University
\qquad
\textsuperscript{2}Snap Research
}

\vspace{-20mm}
\twocolumn[{%
\renewcommand\twocolumn[1][]{#1}%
\maketitle

\begin{center}
    \centering
    \animategraphics[autoplay,loop,width=\textwidth, trim=5 0 5 0, clip]{30}{images/gif_teaser/}{0}{119}
    \vspace{-8mm}
    \captionof{figure}{Our method can synthesize artistic cinemagraphs given text prompts, bringing to life motion effects such as  ``waterfall falling'',  ``river flowing'', and ``turbulent ocean.'' These visual effects may be challenging to depict in a static photograph, but they flourish in the medium of cinemagraphs. To view the teaser images as videos, we recommend using Adobe Acrobat.}
    \label{fig:teaser}
\end{center}
}]
 \maketitle
\ificcvfinal\thispagestyle{empty}\fi

\begin{abstract}
We introduce \camera{Text2Cinemagraph},  a fully automated method for creating cinemagraphs from text descriptions --- an especially challenging task when prompts feature imaginary elements and artistic styles, given the complexity of interpreting the semantics and motions of these images. \camera{We focus on cinemagraphs of fluid elements, such as flowing rivers, and drifting clouds, which exhibit continuous motion and repetitive textures.} Existing single-image animation methods fall short on artistic inputs, and recent text-based video methods frequently introduce temporal inconsistencies, struggling to keep certain regions static. To address these challenges, we propose an idea of synthesizing image \emph{twins} from a single text prompt --- a pair of an artistic image and its pixel-aligned corresponding natural-looking twin. While the artistic image depicts the style and appearance detailed in our text prompt, the realistic counterpart greatly simplifies layout and motion analysis.  Leveraging existing natural image and video datasets, we can accurately segment the realistic image and predict plausible motion given the semantic information. The predicted motion can then be transferred to the artistic image to create the final cinemagraph. Our method outperforms existing approaches in creating cinemagraphs for natural landscapes as well as artistic and other-worldly scenes, as validated by automated metrics and user studies. Finally, we demonstrate two extensions: animating existing paintings and controlling motion directions using text. Please find code and video results on our project \website




\end{abstract}


\section{Introduction}
\label{sec:intro}
Cinemagraphs are captivating visuals where certain elements exhibit repeated, continuous motions while the rest remains static~\cite{cinemagraph}. They offer a unique way to highlight scene dynamics while capturing a specific moment in time~\cite{cohen2006moment}. Since their inception,  cinemagraphs have become popular as short videos and animated GIFs on social media and photo-sharing platforms. They are also prevalent in online newspapers, commercial websites, and virtual meetings. However, creating a cinemagraph is remarkably challenging as it involves capturing videos or images with a camera and using semi-automated methods to produce seamless looping videos. This process often requires considerable user effort~\cite{tompkin2011towards,bai2012selectively}, including capturing suitable footage, stabilizing video frames,  selecting animated and static regions, and specifying motion directions.


In this work, we explore a new research problem of text-based cinemagraph synthesis, significantly reducing the need for data capture and tedious manual efforts. As shown in \reffig{teaser}, our method captures motion effects such as ``water falling'' and ``flowing river''
, which cannot be easily expressed with still photographs and existing text-to-image methods. More importantly, our approach broadens the spectrum of styles and compositions available in cinemagraphs, which allows content creators to specify various artistic styles and describe imaginative visual elements. Our method can synthesize realistic cinemagraphs as well as creative or otherworldly scenes. \camera{Similar to prior works on single-image cinemagraph generation~\cite{holynski2021animating, mahapatra2022controllable, fan2022simulating}, we focus on generating cinemagraphs with predominantly fluid elements like water, smoke, and clouds.} 

This new task presents significant challenges to current methods. One straightforward approach is to use a text-to-image model to generate an artistic image and then animate it. Unfortunately, existing single-image animation methods struggle to predict meaningful motions for artistic inputs as the models are typically trained on real video datasets. 
Curating a large-scale dataset of artistic looping videos is impractical, given the difficulty of producing individual cinemagraphs and the wide range of artistic styles involved.
 Alternatively, text-based video models can be used to generate videos directly. However, our experiments reveal that these methods often introduce noticeable temporal flickering artifacts in static regions and fail to produce desired semi-periodic motions. 

To close the gap between artistic images and animation models developed for real videos, we propose an algorithm based on the concept of \emph{twin} image synthesis. Our method generates two images from a user-provided text prompt -- one artistic and one realistic -- that share the same semantic layout. The artistic image represents the style and appearance of the final output, while the realistic counterpart provides an input that is much easier for current motion prediction models to process. After we predict the motion for the realistic image, we can transfer this information to its artistic counterpart and synthesize the final cinemagraph. Though the realistic image is not displayed as the final output, it serves as a critical intermediate layer that resembles the semantic layout of the artistic image while remaining amenable to existing models. To improve motion prediction, we further leverage additional information from text prompts and semantic segmentation of the realistic image.

Our experiments show that our method outperforms existing single-image animation methods on both artistic and natural images in terms of automated metrics such as Fréchet Video Distance~\cite{unterthiner2018towards}. We also demonstrate that our method synthesizes more visually appealing artistic cinemagraphs compared to existing single-image animation and zero-shot text-to-video methods, according to a user study. We further include an ablation study regarding different algorithmic designs. Finally, we extend our method to two applications: animating existing paintings and text-based control of the motion directions. 

\section{Related Work} 
\label{sec:related}
\myparagraph{Video Looping and Cinemagraph.}  Classic methods for creating cinemagraphs often involve reusing frames from a real video with periodic motions, as demonstrated by Video Texture~\cite{schodl2000video}. This seminal work uses a graph-based formulation to find seamless transitions between frames with similar appearance and motions. 
To accommodate varying motion patterns across different regions, several works~\cite{kwatra2003graphcut,tompkin2011towards, bai2012selectively,joshi2012cliplets} propose separating looping dynamic elements from the static background, which often requires user guidance to mask dynamic regions and stabilize videos. Other works~\cite{liao2013automated,liao2015fast} further allows users to animate each region with different looping periods and handle videos with moving cameras~\cite{sevilla2015smooth}. These techniques have also been applied to process specific visual data such as portraits~\cite{bai2013automatic} and panoramas~\cite{agarwala2005panoramic,he2017gigapixel}. In contrast to these methods, our work does \emph{not} require capturing a periodic real video or additional user annotation. 





\begin{figure*}[t!]
    \centering
    \includegraphics[width=0.99\linewidth]{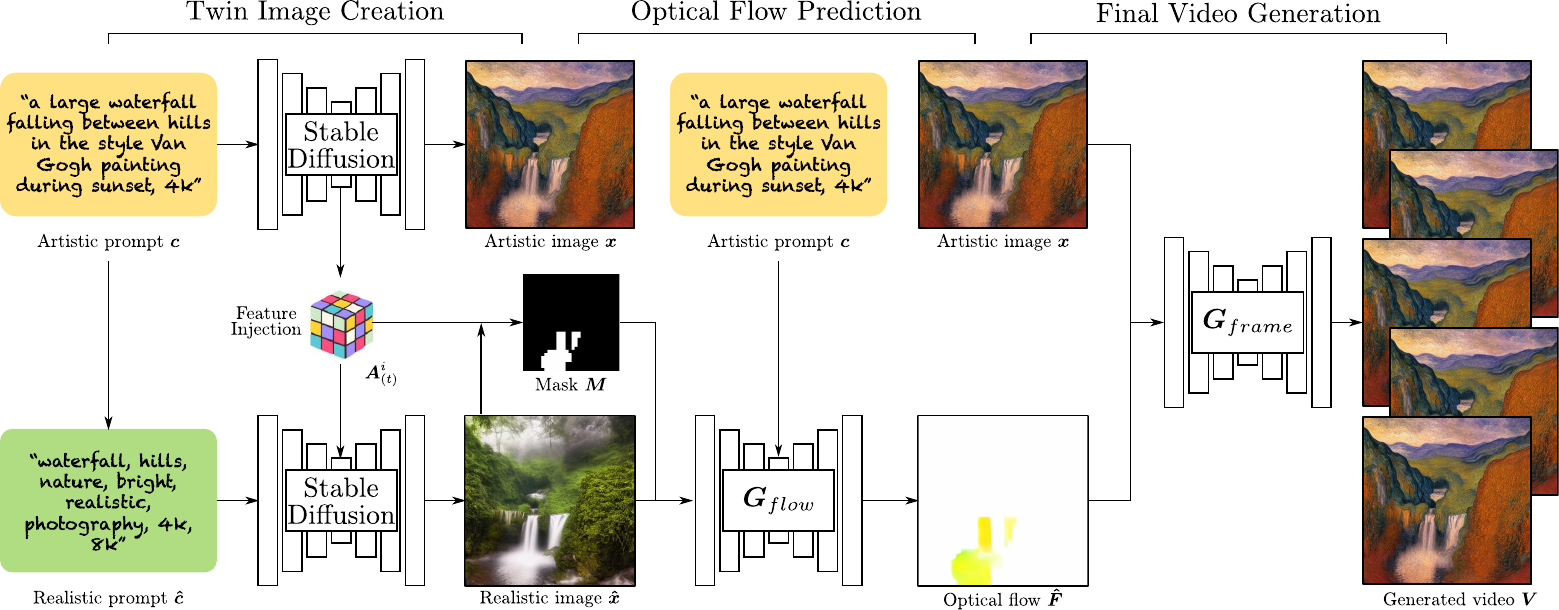}
    \caption{\camera{{\bf Overview.} Given a text prompt $\at$, we generate twin images with Stable Diffusion, an artistic image $\art$ in the style described in the text prompt, and a realistic counterpart $\nat$ using the modified prompt $\nt$. Twin images share a similar semantic layout. We then extract a binary mask $\artmask$ of the moving regions from the Self-Attention maps obtained during the artistic image's generation process. We use the mask and the realistic image to predict the optical flow $\natflow$ with the flow prediction model $\gf$. Since the twin images have a very similar semantic layout, we can use the flow $\natflow$ to animate the artistic image, with our video generator $\gv$.}}
    \lblfig{overview}
\end{figure*}

\myparagraph{Single Image Animation}
Another way of creating cinemagraphs is to start from a single image and add periodic motions. The first notable work in this area was by Chuang et al.~\cite{chuang2005animating}. They manually define different types of motion for different subject classes such as water and leaves. More recent works have employed deep networks to predict motion~\cite{endo2019animating, logacheva2020deeplandscape, holynski2021animating, mahapatra2022controllable, fan2022simulating,li20233d,bertiche2023blowing}. Animating Landscape~\cite{endo2019animating} predicts motion autoregressively and generates the results video using backward warping. Holynski et al.~\cite{holynski2021animating} instead predicts a single optical flow that describes the motion between consecutive frames and uses forward warping to reduce stretching artifacts. Mahapatra et al.~\cite{mahapatra2022controllable} and Fan et al.~\cite{fan2022simulating} offer controllable cinemagraph generation methods given user-provided masks and direction hints. \camera{DeepLandscape~\cite{logacheva2020deeplandscape} adopts a StyleGAN model~\cite{karras2019style} to learn motion with random frame pairs from a video.} Although all these methods can animate real images, they struggle with artistic scenes. 
In contrast, our method can support both natural scenes and imaginative, artistic styles. \camera{Recently, EndlessLoops~\cite{halperin2021endless} provide a controllable and non-deep learning-based method to animate a single image that can work on a wide range of scenes, but requires extensive user input arrows and precise masks.} Our method is the first text-based cinemagraph synthesis approach, significantly reducing the manual efforts involved in image capture and user annotations.

\myparagraph{Text-to-image Synthesis.} Text-based interfaces have gained widespread usage in image synthesis with deep generative models,  such as GANs~\cite{patashnik2021styleclip,gal2022stylegan,kang2023gigagan,Sauer2023ICML}, Diffusion models~\cite{saharia2022photorealistic,ramesh2022hierarchical,gafni2022make}, autoregressive models~\cite{yu2022scaling}, and their hybrids~\cite{rombach2022high,crowson2022vqgan,esser2021taming}. These approaches offer two major advantages.  First, the text provides a \camera{widely accessible medium to many users}. Second, recent text-based models demonstrate the ability to synthesize high-quality images with diverse styles and content. We aim to harness these advantages in our cinemagraph synthesis method. However, visualizing motion effects like ``falling water'' presents inherent challenges with still photographs. To overcome this limitation, our work depicts these visual effects with cinemagraphs. 

More recently, several works have focused on text-based image editing, aiming to preserve the structure and content of the input image while incorporating desired changes specified by a text prompt~\cite{brooks2022instructpix2pix,saharia2022palette,wang2022pretraining,zhang2023adding,tumanyan2022plug,kawar2022imagic,parmar2023zero}. We adopt some of them to bridge the gap between real video training datasets and artistic text prompt.  

\myparagraph{Text-to-video Models.} Another line of work attempts to directly synthesize videos based on text descriptions. Similar to the image domain, the most prominent directions are autoregressive~\cite{hong2022cogvideo, villegas2023phenaki} and diffusion models~\cite {ho2022imagen,singer2022make,ge2023preserve,blattmann2023videoldm,videocrater}. ImagenVideo~\cite{ho2022imagen} and Make-a-Video~\cite{singer2022make} propose using a cascade of diffusion models for different resolutions. Alternatively, several methods~\cite {blattmann2023videoldm, videocrater} use the pretrained Stable Diffusion~\cite{stablediffusion}. 
All these works require a significant amount of data and computational resources for training. A notable exception is Text2Video~\cite{khachatryan2023text2video}, which uses only a pretrained Stable Diffusion model, without any additional fine-tuning to generate a video. Nonetheless, none of these methods explicitly model motion. As a result, our experiments demonstrate that the temporal consistency of their generated videos, a key aspect of plausible cinemagraphs, falls short compared to our method. 



\section{Method} 
\label{sec:method}

Given a user-provided artistic text prompt $\at$, describing an artistic or imaginative scene, along with a user-specified region name that the user wants to animate (e.g., waterfall), our goal is to generate a cinemagraph that faithfully reflects the text description.

As demonstrated in \refsec{experiments}, existing text-to-video models often fail to generate visually pleasing cinemagraphs due to several factors. First, these models have \emph{not} been trained on cinemagraph datasets. Curating such datasets becomes a chicken-and-egg problem, as creating a \emph{single} artistic cinemagraph usually requires the expertise of a professional artist with existing tools. Second, text-to-video models frequently generate moving-camera effects and struggle to maintain temporal coherence in predominantly static scenes.

Instead, one intuitive method involves first generating an artistic image $\art$ using Stable Diffisuion~\cite{rombach2022high} in the same style as described by the text prompt. This is followed by predicting plausible motion corresponding to the image, and finally animating the image. Unfortunately, a na\"ive implementation of this idea fails in practice due to several challenges. 

The first critical challenge lies in predicting plausible motion for artistic images or imaginative content. It is a daunting task, as motion prediction models are typically trained only on real video datasets. 
To tackle this, we propose generating a twin realistic image $\nat$, sharing the same semantic structure as the artistic image $\art$, but with a natural photography style. 
Conveniently, we can automatically derive the realistic image $\nat$  by leveraging intermediate diffusion model features produced during the artistic image generation process. We elaborate on this in Section~\ref{sec:twin}.

The second challenge is that single-image motion prediction remains hard even for natural images. Recent methods~\cite{holynski2021animating,endo2019animating} often hallucinate movements for stationary regions (like rocks and hills), leading to noticeable artifacts. To address this challenge, we propose using a binary mask to direct the prediction model toward the dynamic regions.  Yet again, the question arises: ``How do we obtain such a mask for an artistic image?''. Once more, the twin realistic image comes to the rescue. We leverage a pretrained segmentation model, trained on real images, ODISE~\cite{xu2023open} and the user-specified region name, to predict a binary mask for realistic image $\nat$. 
To reduce the misalignment between region boundaries between twin images $\nat$ and $\art$, we further refine this mask using intermediate diffusion model features. The whole mask and motion prediction algorithm is described in Section~\ref{sec:mask}.


Finally, in Section~\ref{sec:cinemagraph}, we transfer the predicted motion from the realistic image to the artistic image and generate a coherent cinemagraph. \reffig{overview} illustrates our method. 

\begin{figure}[t!]
    \centering
    \includegraphics[width=\linewidth]{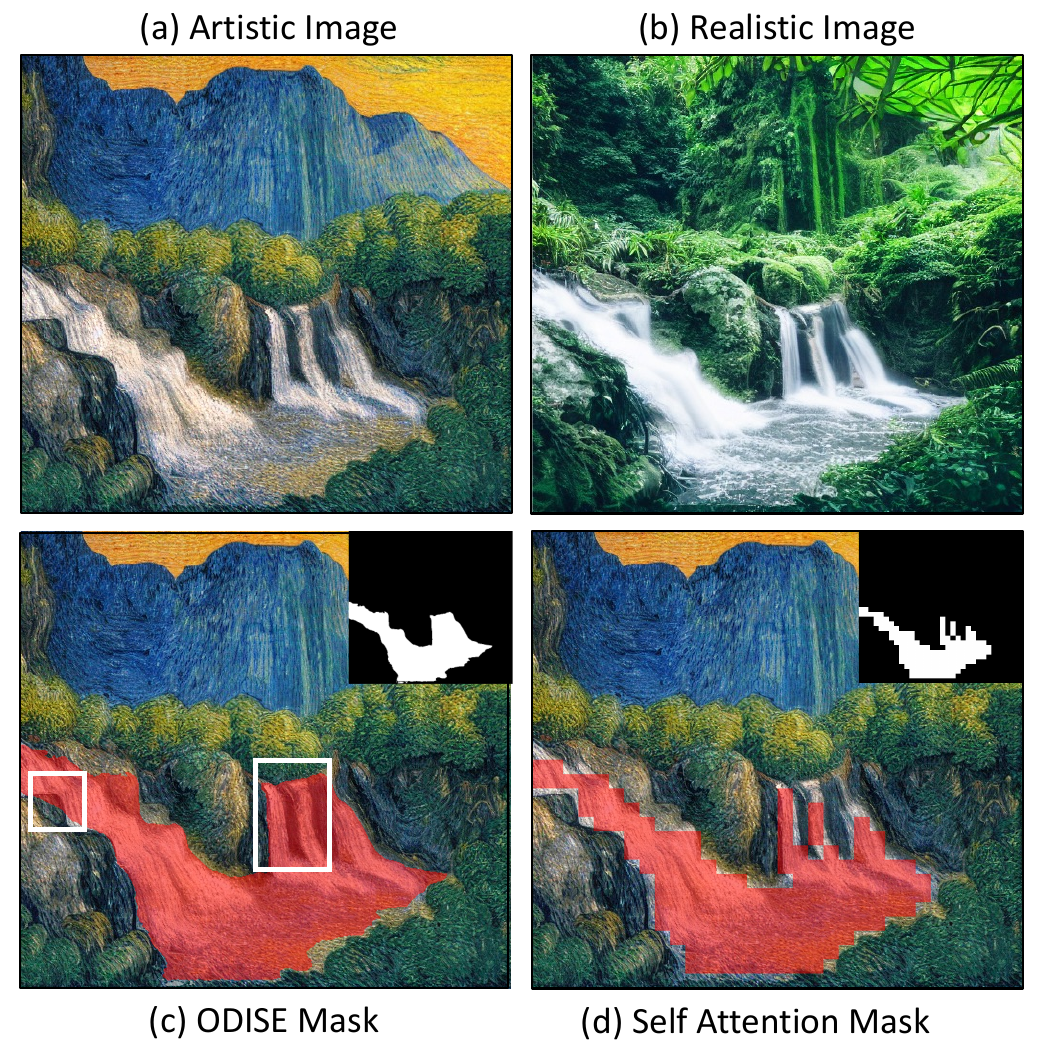}
    \caption{{\bf \camera{ODISE vs. Self-Attention Mask}.} We aim to derive a mask, given twin images: (a) the generated artistic image and (b) its realistic counterpart. 
    Directly applying a segmentation model (e.g., ODISE~\cite{xu2023open}) to the realistic image might introduce segmentation errors. For example, in the ODISE-generated mask on the artistic image (c), some regions in the mask overlap with hills, which can result in the movement of the hills. 
   We use the diffusion model's self-attention maps to further refine the mask (d).  
    }
    \label{fig:mask}
    
\end{figure}

\subsection{Twin Image Generation}
\label{sec:twin}

We start by generating an artistic image, $\art$, that corresponds to the input text prompt using a pretrained Stable Diffusion model. We assume that this image can have some unnatural style, such as "pixel art" or "Monet painting".  
As mentioned before, for our motion prediction network to work, we need a twin version of the artistic image, denoted as $\nat$, with a similar semantic layout.


To generate the natural image, we need a corresponding natural-styled prompt, $\nt$, distinct from the user-provided artistic prompt $\at$. Rather than expecting users to provide  $\nt$ separately, we automatically generate the prompt $\nt$ that contains all the \camera{common nouns ('NN', 'NNS')} from the input prompt $\at$, using the NLTK natural language toolkit~\cite{nltk} \camera{corresponding to COCO-stuff~\cite{caesar2018coco} or DTDB~\cite{hadji2018new} classes, along with their synonyms} and append \textit{``nature, bright, realistic, photography, 4k, 8k''} at the end. For example, given the $\at$ as \camera{\textit{``a large waterfall falling between hills in the style Van Gogh painting during sunset, 4k''}, its corresponding $\nt$ would be \textit{``waterfall, hills, nature, bright, realistic, photography, 4k, 8k''}}.


Simply using the natural-styled prompt $\nt$ along with the same initial seed that was used to generate the artistic image does not guarantee that the twin images will always share the same semantic layout. In fact, this is rarely the case.  Following recent works~\cite{parmar2023zero, tumanyan2022plug, hertz2022prompt}, we can further use the diffusion model's internal features to enforce structural correspondence between the two images. 

Similar to Plug-and-Play~\cite{tumanyan2022plug}, we observe that self-attention maps $\bs{A}_t$, at any timestep $t$ in the denoising process in Stable Diffusion, control the spatial structure of the resulting image. Thus, during the generation of the artistic image $\art$, we store the intermediate self-attention maps $\bs{A}_t$ for all timesteps $t$, As noted by Plug-and-Play~\cite{tumanyan2022plug}, the structural alignment can be made even stronger by injecting the output residual block features $\feature$. We store the $4$-th layer in the output blocks. 
\begin{equation}
    \arttminus, \atten, \feature = \bs{\epsilon_\theta}(\artt, \at, t), 
\end{equation}
where $ \bs{\epsilon_\theta}$ is a standard denoising UNet.  
To enforce structural similarity when generating the realistic image $\nat$, we inject self-attention maps $\bs{A}^i_t$ and residual block features $\feature$ into the UNet module. 
\begin{equation}
     \nattminus = \bs{\hat{\epsilon}_\theta}(\natt, \nt, t\;;\;  \atten, \feature),
\end{equation}
where $\artt$, $\natt$ are noisy images at timestep $t$ corresponding to the artistic and realistic image, respectively, and $\bs{\hat{\epsilon}_\theta}$ is the modified UNet, which also takes injected features as input. 
Note that Stable Diffusion operates in the latent space, but for notation brevity, we use $\art$ and $\nat$ notation instead of introducing new notations for latents.



By leveraging the stored attention maps and residual block features, we ensure that the generated realistic image $\nat$ maintains a similar semantic layout to its artistic counterpart $\art$. This enables us to establish a meaningful correspondence between the twin images, which is crucial for subsequent steps.

\subsection{Mask-guided Flow Prediction}
\label{sec:mask}

As mentioned previously, predicting plausible optical flow requires an input binary mask that defines the regions to animate in addition to an image. We use a pretrained open set panoptic segmentation module, such as ODISE \cite{xu2023open}, to generate the mask. For example, for a prompt \textit{``a large river flowing in front of a mountain in the style of starry nights painting''}, and a user-specified region name \textit{`river'}, we would use the segmentation map corresponding to \textit{`river'} as the binary mask. 
However, reliable masks can only be predicted for the realistic image since the segmentation network is trained on real-world data.
Although the images are twins, we cannot use the mask obtained for the realistic image directly for predicting flow as there may be inconsistencies at the boundaries of semantic regions between the two images. In \reffig{mask}, we illustrate this issue with an example. This will result in motion prediction at the boundaries of static regions in the realistic image creating noticeable artifacts. 

\begin{figure}[t!]
    \centering
    \includegraphics[width=\linewidth]{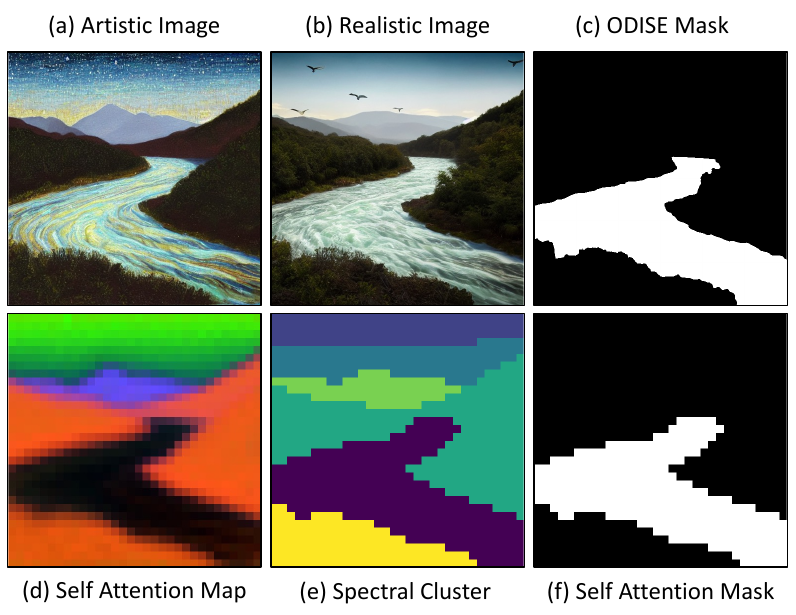}
    \caption{
    {\bf \camera{Mask generation pipeline.}}
   \camera{To generate our final binary mask $\artmask$, we first generate a realistic twin (b) corresponding to the artistic image (a). Using a pre-trained open set panoptic segmentation module, ODISE \cite{xu2023open}, and a user-specified region name \textit{`river'}, we generate a binary segmentation mask (c). 
    In (d), we use PCA~\cite{pca} to visualize the self-attention maps averaged across different timesteps. 
    Finally, we apply spectral clustering~\cite{ng2001spectral} on the average self-attention maps (e) and select regions with high Intersection-over-Union (IOU) scores with respect to the ODISE mask (c). 
    }
    }
    \label{fig:mask_pipeline}
    
\end{figure}

To this end, we use the self-attention maps $\atten$ to generate the mask. We first average $\atten$, across different timesteps $t$ and obtain average self-attention map $\attenavg$. We only use the self-attention maps after a certain number of steps, as earlier maps are noisy. We then cluster the pixels by applying spectral clustering~\cite{ng2001spectral} to the average self-attention map. We then predict a binary mask $\natmask$ for the realistic image $\nat$ \camera{using ODISE}. Finally, we use the predicted mask $\natmask$ to select which clusters to retain in $\attenavg$ based on the Intersection-over-Union (IOU) percentage between the two to obtain the final binary mask $\artmask$. \camera{\reffig{mask_pipeline} shows the different stages in our mask generation method.} 

\camera{Unlike concurrent works~\cite{patashnik2023localizing, ge2023expressive}, we use ODISE mask instead of cross-attention map (for \textit{`river'} token) to select self-attention map regions. 
We found that ODISE produces more precise masks for natural scenes (like \camera{ocean}, rivers, etc.). For example, for the prompt, ``a view of a rocky beach on the shore of the ocean’’, the cross-attention map for the 'ocean' token highlights both the ocean and the surrounding rocky beach. More details are provided in the supplementary material.}
Following this, we predict the optical flow ($\natflow$) on the realistic image $\nat$, using a generator $\gf$ conditioned on $\artmask$.
In addition, we condition the flow prediction on the CLIP \cite{radford2021learning} embedding input text prompt $\at$,
\begin{equation}
    \natflow = \gf(\nat, \at, \artmask),  
\end{equation}
where we compute CLIP embedding given the text prompt $\at$ before feeding to the flow prediction network. We condition the optical flow prediction model with the text through cross-attention layers. We use the same spatial transformer blocks as in Stable Diffusion, but omit the self-attention layers for text conditioning. Due to the limited dataset size, we initialize the spatial transformer blocks with weights of Stable Diffusion and freeze the key and query weight matrices. We train the flow prediction network $\gf$ with End-Point-Error (EPE), conditional GAN loss~\cite{isola2017image}, and feature matching loss~\cite{wang2018pix2pixHD}.  Please refer to our supplement for more training details. We hypothesize that text inherently contains class information, like a \textit{`waterfall'} or \textit{`river'}, which can be useful to determine the natural direction in the predicted flow. 

\subsection{Flow-guided Video Generation}
\label{sec:cinemagraph}


Although in \refsec{mask}, we predict flow $\natflow$ for the realistic image $\nat$, we can use this flow to animate the artistic image $\art$, as they both share a similar semantic layout. 
We, therefore, define $\artflow = \natflow$, assuming that we can directly transfer the motion of natural image $\natflow$ to its artistic counterpart $\artflow$. 
   
Given an accurate optical flow $\artflow$, we can animate $\art$ using a method similar to Holynski et al.~\cite{holynski2021animating} \camera{for animating a single image, wherein each frame in the cinemagraph is generated separately}. Since \camera{cinemagraphs have the property of looping in time}, the artistic image $\art$ serves as the first and the last frame. Considering a total of $N$ frames in the generated cinemagraph, we generate a frame $n \in [0, N]$ by symmetric splitting, \camera{introduced in Holynski et al.~\cite{holynski2021animating}}. For this, we need the cumulative flow $\bs{F}_{0\to n}$, and $\bs{F}_{N\to N-n}$ in forward and backward directions respectively, where $\bs{F}_{0\to n}$, and $\bs{F}_{N\to N-n}$ are optical flows between frame $(0, n)$ and $(N, N-n)$. The predicted optical flow $\artflow$ however defines the flow between any two consecutive frames $n, n+1$. To get cumulative flows in the forward and backward direction, we perform Euler integration of predicted optical flow, $\artflow$, and its reverse flow -$\artflow$, for $n$ and $N-n$ times respectively,

\begin{equation}
\begin{split}
     \bs{F}_{0 \to n}(\pixel) = \bs{F}_{0 \to n-1}(\pixel) + \artflow(\pixel+\bs{F}_{0 \to n-1}(\pixel)) \\
     \bs{F}_{N \to n}(x) = \bs{F}_{N \to N-n+1}(\pixel) - \artflow(\pixel+\bs{F}_{N \to N-n+1}(\pixel)),
\end{split}
\end{equation}
where $\pixel$ is the 2D pixel coordinate in the flow $\artflow$.
$\gv$ is trained to predict an intermediate frame $\bs{v}_{n}$, given the cumulative flows, $\bs{F}_{0\to n}$ and $\bs{F}_{N\to N-n}$, the first frame and the last frame. We perform symmetric splatting in the feature space of an encoder part of the generator, at multiple feature resolutions, similar to Mahapatra et al.\cite{mahapatra2022controllable}, and then use the decoder part to generate the RGB image. We train this model on pairs of cumulative ground-truth flows $\bs{F}^{gt}_{0 \to n-1} \text{and } \bs{F}^{gt}_{N \to N-n}$, the first frame $\bs{v}_{0}$, and the last frame $\bs{v}_{N}$ of ground-truth natural videos to predict an intermediate frame $\bs{v}_{n}$,
\begin{equation}
    \bs{v}_n = \gv(\bs{F}^{gt}_{0 \to n-1}, \bs{F}^{gt}_{N \to N-n}, \bs{v}_{0}, \bs{v}_{N}).
\end{equation}
Our training objective involves $L1$ reconstruction loss, conditional GAN loss~\cite{isola2017image}, feature matching loss~\cite{wang2018pix2pixHD} and VGG-based perceptual  loss~\cite{zhang2018perceptual}. 

Since at test time we want to generate looping videos, we use the artistic image $\art$ as the first and the last frame,
\begin{equation}
    \bs{v}_{n} = \gv(\bs{F}_{0 \to n-1}, \bs{F}_{N \to N-n}, \art, \art).
\end{equation}
Surprisingly, although this model is trained on a dataset of real-domain videos, we can use it to animate the artistic image without any further modification as it essentially performs impainting of the small holes in feature space generated during symmetric splatting, with repetitive surrounding textures.

\section{Experiments} \label{sec:experiments}

In this section, we extensively compare our method against recent methods, 
both qualitatively and quantitatively, on real-world single-image animation (termed real domain) and text-to-cinemagraph for the artistic text prompts (termed artistic domain).

\myparagraph{Dataset.} Our dataset contains two domains: (1) \emph{Real Domain}: We train our optical flow prediction and animation models, $\bs{G}_{flow}$ and $\bs{G}_{frame}$, on the dataset provided by Holynski et al~\cite{holynski2021animating}. This dataset contains real-life videos of waterfall, lake, river, and ocean scenes with ground-truth average optical flow calculated for each video.  Each video consists of 60 frames with a resolution of $720 \times 1280$. The training set contains $4750$ videos, and the test set contains $162$ videos. 
We use BLIP~\cite{li2022blip} to generate captions based on the first frame of each video. 
(2) \emph{Artistic Domain}: Since there is no existing dataset for text-to-cinemagraph, we generate $20$ different captions corresponding to waterfalls, rivers, lakes, clouds, sea, and ocean using different artistic styles with 5-6 seeds each, generating $102$ artistic images. Our data will be available upon publication. 

\begin{figure*}[t!]
    \centering
    \includegraphics[width=\linewidth]{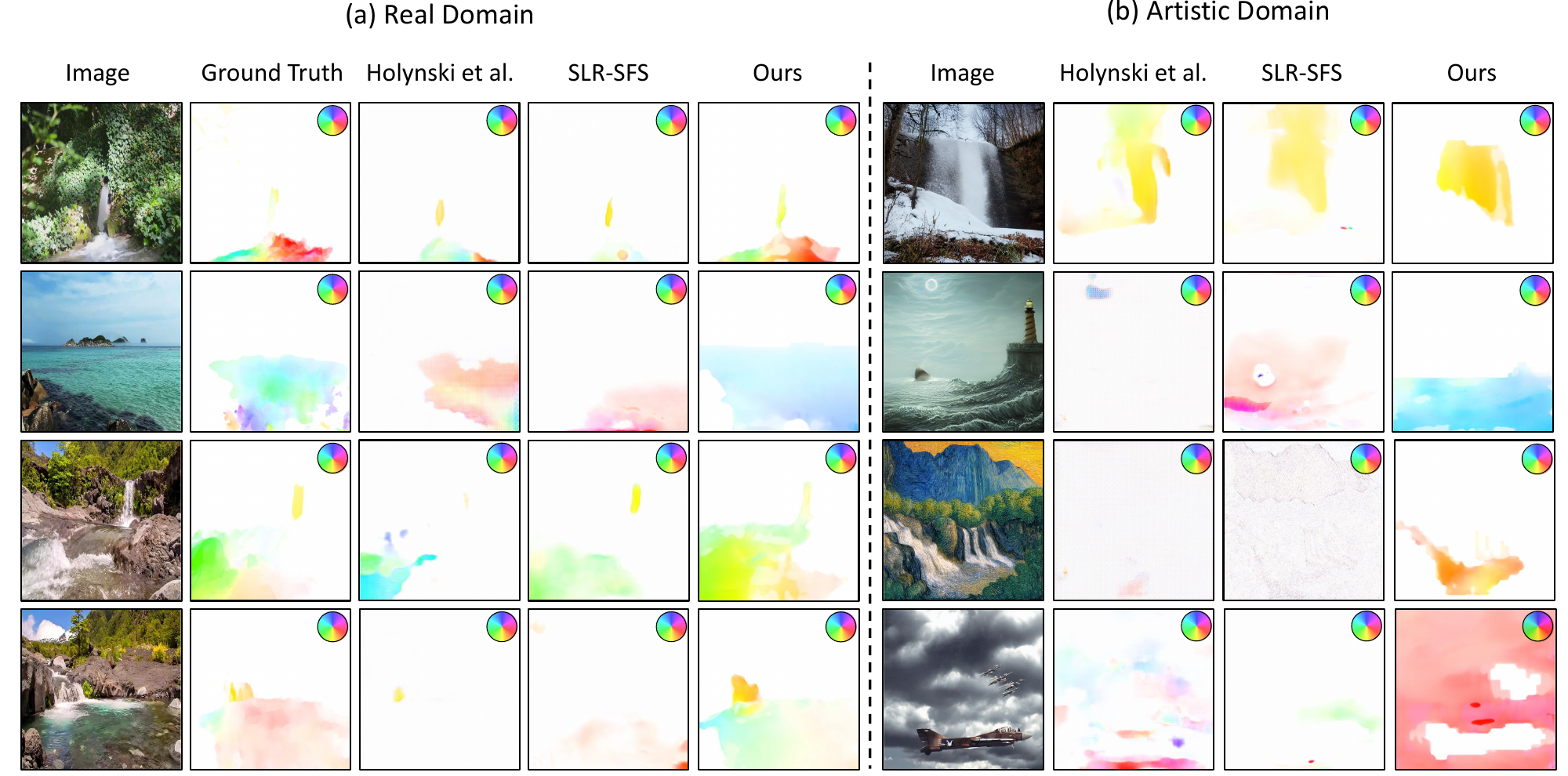}
    \caption{{\bf Visual comparisons for optical flow prediction.}  We compare our method with two single-image animation methods Holynski et al.~\cite{holynski2021animating} and SLR-SFS~\cite{fan2022simulating} on real videos (a) and artistic images (b). For real videos, we also show the ground truth optical flow averaged across all frames. Overall, our method predicts more plausible motions that align better with  target regions. Please see our supplement for more video comparisons with more baselines. }
    \label{fig:comparisons}
\end{figure*}

\myparagraph{Training Details.} For the flow prediction network $\bs{G}_{flow}$, we adopt a UNet backbone~\cite{ronneberger2015u}  with SPADE ResBlocks~\cite{park2019semantic} for image and mask conditioning and cross-attention layers for text conditioning~\cite{mansimov2016generating}. For the animation network $\bs{G}_{frame}$, we also use a UNet backbone. 
In our experiments, we tried using both ODISE~\cite{xu2023open} and Segment Anything~\cite{kirillov2023segany}, and observed that Segment Anything produces a large number of holes in the segmentation masks, while ODISE produced much more consistent masks for our type of scene.
Similar to Mahapatra et al.~\cite{mahapatra2022controllable}, we perform symmetric splatting across different hierarchical features of UNet. To ensure stable training, we first train two networks $\bs{G}_{flow}$ and $\bs{G}_{frame}$, separately, each for 200 epochs. We then train both models end-to-end for 50 epochs. Since the test videos in the real video dataset contain 60 frames, we generate videos of 60 frames duration. For artistic prompts, we generate videos of $120$ frames. More details are provided in the supplementary material.
\myparagraph{Evaluation Metrics.} We compare the quality of generated cinemagraphs on real-domain data using Fréchet Video Distance (FVD)~\cite{unterthiner2018towards}, a commonly-used metric used in video generation~\cite{ho2022imagen,skorokhodov2022stylegan}. We adopt two FVD variants, as suggested by Skorokhodov et al.~\cite{skorokhodov2022stylegan}: (1) FVD$_{16}$  calculates the FVD score on $16$ frames sampled at a factor of 3, and (2) FVD$_{60}$ computes the FVD score using all generated frames. 

\myparagraph{User Study.} We also conduct a user study to assess the quality of generated cinemagraphs on both real and artistic domains. We ask the Amazon MTurk participants to choose which animation they prefer. For real videos, we ask them to make choices based on two criteria: (1) which video has more natural movement, according to what occurs in the real world, and (2) looks better visually. For artistic images, we keep (1) and (2), and add a third criterion (3) the video describes what is written in the text accurately.  \camera{We perform paired tests, where the users are asked to compare two videos, one generated by our method and the other by one of the baseline or our ablation method. Each paired comparison is annotated by $5$ annotators. More details about the User Study are provided in the supplementary material.}

For artistic cinemagraph generation, we rely solely on the user preference study, as we do not possess ground-truth data. 

\begin{table}[t!]
    \centering
    \resizebox{0.75\linewidth}{!}{
    \begin{tabular}{l c c}
        \toprule
        \textbf{Method} & \textbf{FVD$_{16}$} & \textbf{FVD$_{60}$}
        \\
        \cmidrule(lr){1-3}
        Animating Lanscape~\cite{endo2019animating} &  $1122.87$ & $1276.43$ \\
        Holynski et al.~\cite{holynski2021animating} & $787.03$ & $946.28$ \\
        SLR-SFS~\cite{fan2022simulating} & $773.04$ & $909.88$ \\
        
        \cmidrule(lr){1-3}
        Ours (w/o text and mask) & $736.93$ & $981.54$ \\
        Ours (w/o mask) & $735.75$ & $936.53$ \\
        Ours (w/o text) & $662.64$ & $695.5$ \\
        Ours (full) & $\mathbf{659.48}$ & $\mathbf{689.08}$ \\
        \bottomrule 
    \end{tabular}
    }
    \caption{{\bf Quantitative comparisons regarding video quality.} Here we compare the generated videos with ground truth videos regarding Fréchet Video Distance (FVD) on real video datasets. 
    }
    \label{tab:fvd}
\end{table}

\begin{figure*}[t!]
    \centering
    \includegraphics[width=\linewidth]{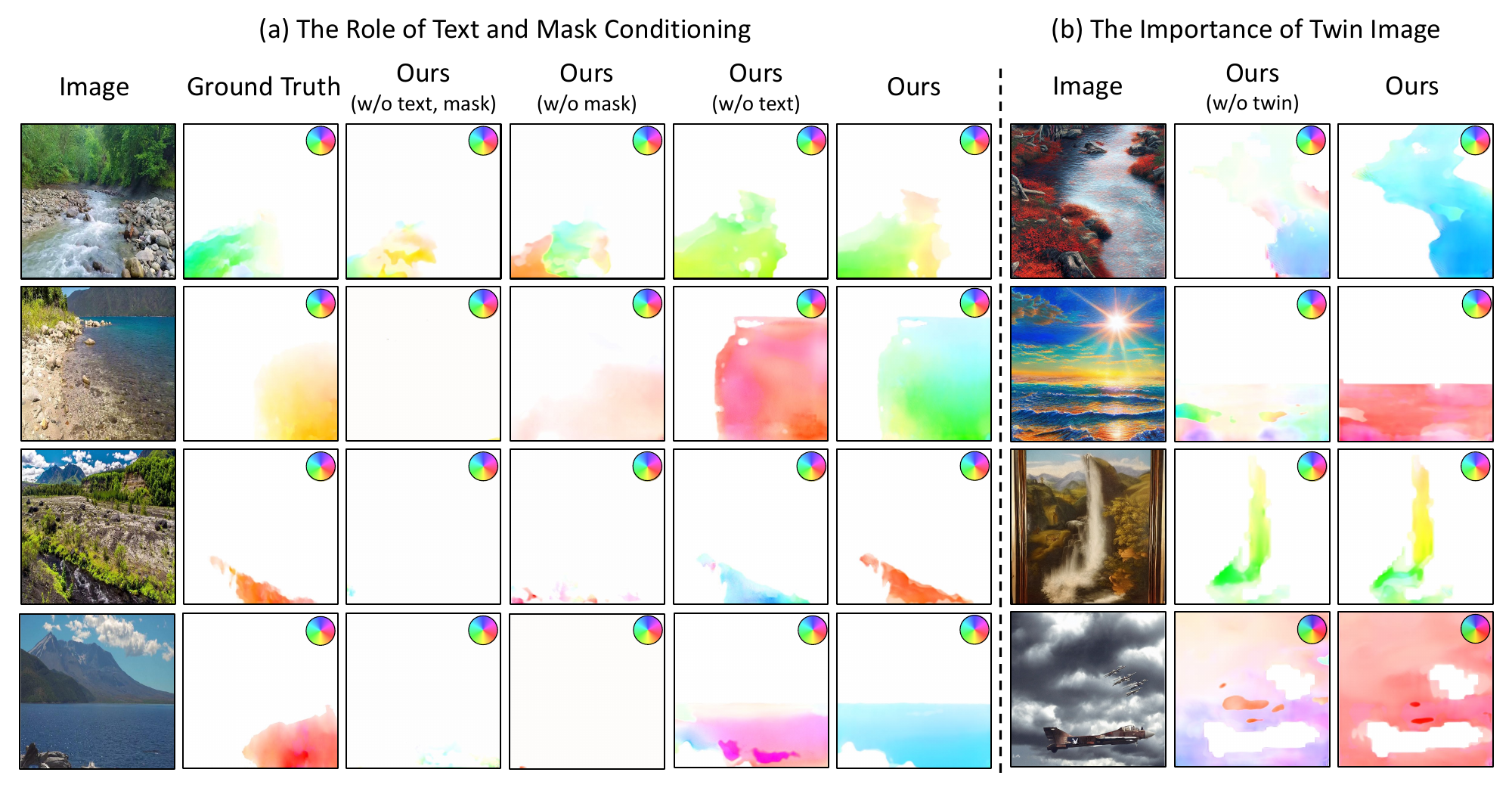}
    \caption{{\bf Ablation Study (flow prediction).}  (a) The role of text and mask conditioning:  we compare our flow prediction method with three variants: (1) w/o mask, (2) w/o text, and (3) w/o mask and text conditioning. Both text embedding vectors and semantic masks contribute to predicting more plausible flows. For the 1st row, our predicted flow covers more water regions than ground truth flow. For the 2nd row, our predicted flow is physically plausible, even when it differs from ground truth flows. (b) The role of twin image synthesis: Directly predicting flows given artistic images will incur significant errors.  
    }
    \label{fig:ablation}
\end{figure*}

\begin{table*}[t!]
    \centering
    \resizebox{0.95\linewidth}{!}{
    \begin{tabular}{l c c c c c c}
        \toprule
        \textbf{Domain} & Animating Landscape~\cite{endo2019animating} & Holynski et al.~\cite{holynski2021animating} & SLR-SFS~\cite{fan2022simulating} & CogVideo~\cite{hong2022cogvideo} & Text2Video-Zero~\cite{khachatryan2023text2video} & VideoCrafter~\cite{videocrater}
        \\
        \cmidrule(lr){2-7}
        Real Domain & $95.26 \pm 1.42 \%$ & $65.75 \pm 3.4 \%$ & $66.18 \pm 3.17 \%$ & - &  - & - \\
        Artistic Domain & $94.35 \pm 6.37 \%$ & $76.24 \pm 5.41 \%$ & $85.82 \pm 6.99 \%$ & $84.01 \pm 7.57 \%$ & $82.13 \pm 7.96 \%$ & $84.12 \pm 7.58\%$ \\
        \bottomrule 
    \end{tabular}
    }
    \caption{{\bf User pereference study}. 
   The numbers indicate the percentage of participants who prefer our results over those of the baselines, given the same text prompt. The criteria include visual quality, the naturalness of movement, and text-video alignment (only for artistic images).  We compare our method with single-image animation methods~\cite{endo2019animating,holynski2021animating,fan2022simulating} and text-based video models~\cite{hong2022cogvideo,khachatryan2023text2video,videocrater}. Most participants favor our results. 
    }
    \label{tab:user_study}
\end{table*}

\subsection{Baselines.} 
\myparagraph{Single Image Animation.} (1) Animating Landscape~\cite{endo2019animating} predicts time-varying optical flow and generate the next frame using backward warping, autoregressively 
They post-process the generated video to create a looping effect.  
(2) Holynski et al.~\cite{holynski2021animating} predict a constant optical flow from a given image and use this optical flow to generate a video using Euler integration and symmetric splatting. Their method inherently generates looping videos. (3) SLR-SFS~\cite{fan2022simulating} is designed for controllable animation that requires user-specified flow hints and masks. In our case, we repurpose their method for single-image animation without user controls for a fair comparison. 

For text-to-cinemagraph, given a text prompt, we first generate an output image using Stable Diffusion~\cite{rombach2022high}. We then animate the generated image using the above methods.


\myparagraph{Text-to-Video.} (1) CogVideo \cite{hong2022cogvideo} is based on a transformer text-to-image model CogView~\cite{ding2021cogview}. (2) Text2Video-Zero \cite{khachatryan2023text2video} generates videos from text using a pretrained Stable Diffusion directly. (3) VideoCrafter~\cite{videocrater} train a text-to-video diffusion model to generate a video from text. For our comparison, we use pretrained models provided by the authors. 
The above models are recently released open-source models, while other text-to-video models~\cite{ho2022imagen,singer2022make,zhou2022magicvideo} are close-sourced.  
Additionally, as the videos produced by these models are non-looping, we implement a postprocessing method~\cite{endo2019animating} to make them loop.


\subsection{Real Domain Results}

In this section,  we compare the quality of predicted optical flow and generated cinemagraphs from a single image on a real video dataset~\cite{holynski2021animating}. 

\myparagraph{Qualitative Comparison.} As shown in \reffig{comparisons}a, our method predicts more plausible flows than baselines. Due to the lack of mask conditioning, both Holynski et al.~\cite{holynski2021animating} and SLR-SFS~\cite{fan2022simulating} predict flow in static regions and ignore many parts in the dynamic regions.  Holynski et al.~\cite{holynski2021animating}'s flow visualizations also exhibit checkerboard artifacts. In contrast, our method  predicts flows spanning the entire dynamic regions (like \textit{`river'}). Although the directions can sometimes vary from the ground truth, the flows remain physically plausible. For instance,  a river can move from left to right or right to left, as shown in \reffig{ablation}a (row 2). Additionally, as shown in \reffig{comparisons}a (row 3) and \reffig{ablation}a (rows 1 and 2), our flow sometimes covers more regions than the ground truth due to the precise masks predicted by ODISE, while the ground-truth flow is derived from the real video using the RAFT optical flow algorithm~\cite{teed2020raft}, which may neglect flow in regions of less movement. Please check out all the video comparison results on the project \website.


\myparagraph{Quantitative Comparison.} 
We further compare the quality of generated cinemagraphs against ground-truth videos with FVD$_{16}$ and FVD$_{60}$ metrics~\cite{unterthiner2018towards}. From Table \ref{tab:fvd}, we see that our method achieves significantly lower FVD scores compared to baselines. 
Compared to the baselines, our generated cinemagraphs more closely match the data distribution and fidelity of ground-truth videos. 
Our user study, in Table \ref{tab:user_study} (row $1$) further suggests that users prefer the visual quality of our results by a large margin compared to the baselines. 
\begin{figure}
    \centering	
    \animategraphics[autoplay,loop,width=0.49\textwidth, trim=25 0 25 0, clip]{30}{images/gif_painting/}{0}{119}
    \caption{\textbf{Real Painting Cinemagraphs.} Examples of cinemagraph generation from two real paintings. To view this figure as a video, we recommend using Adobe Acrobat.}
    \label{fig:painting}
\end{figure}

\subsection{Artistic Domain Results}
\myparagraph{Qualitative Comparison.} \reffig{comparisons}b shows that our predicted flows are cleaner and focus on the desired regions compared to the baselines, which predict inaccurate  flows and introduce more artifacts. CogVideo~\cite{hong2022cogvideo} struggles to capture all the details in the text prompt. Similarly, VideoCrafter~\cite{videocrater} fails to embody the style mentioned in the text prompts as it has not been trained with these artistic captions. Text2Video-Zero~\cite{khachatryan2023text2video} fails to preserve temporal consistency across frames, though it can capture the details with the pretrained Stable Diffusion. Additionally, all these text-to-video methods generate fewer frames, and animating them by postprocessing introduces cross-fade artifacts. Please see more video results of our method on the project \website. 

\myparagraph{Quantitative Comparison.} Due to the lack of ground-truth artistic cinemagraphs, we compare our method with baselines with a user study. Most users prefer our results to baselines, see \reftbl{user_study} (row $2$). 


\subsection{Ablation Study}
\myparagraph{Text and Mask Conditioning.} We evaluate the design choices of our flow prediction method and compare them against the following variants: (1) w/o mask, (2) w/o text, and (3) w/o mask and text conditioning. As shown in \camera{\reftbl{fvd}}, our method performs the best regarding both FVD scores and slightly outperforms Ours (w/o text). We hypothesize that the text description contains class information (like \textit{`waterfall'}), making it easier for the model to generate plausible flow corresponding to each class category. Ours (w/o mask) and Ours (w/o mask and text) perform the worst, underscoring the mask's critical role.  These findings are corroborated by the qualitative comparisons (\reffig{ablation}). In our user study, $64.11 \pm 3.43 \%$, $63.38 \pm 3.32 \%$, and $58.26 \pm 3.27 \%$ of participants prefer our method to Ours (w/o text and mask), Ours (w/o mask), and Ours (w/o text).

\begin{figure}[t!]
    \centering
    \includegraphics[width=\linewidth]{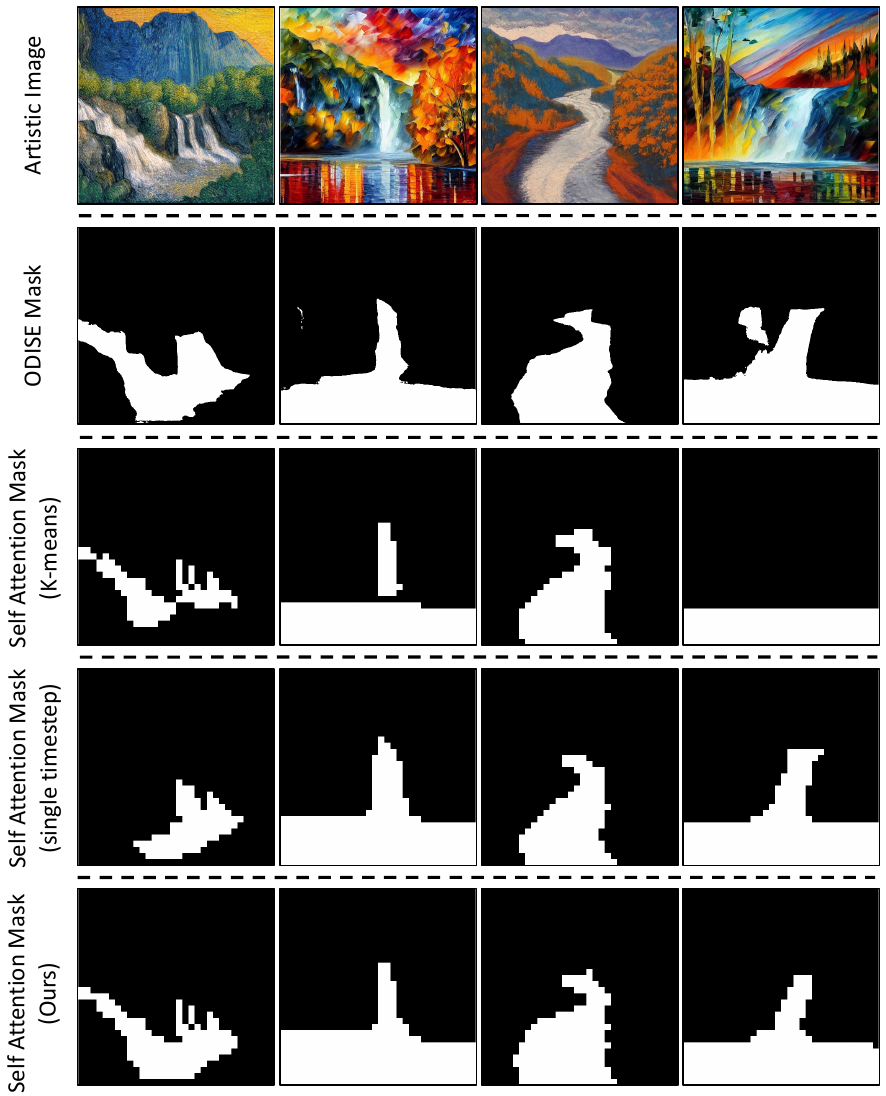}
    \caption{{\bf \camera{Ablation Study (mask generation).}} 
    \camera{Mask generation with different design choices: we compare the quality of masks generated from our method (a) with alternative design choices like (1) using K-Means~\cite{lloyd1982least} clustering instead of spectral clustering~\cite{ng2001spectral}, (2) using self-attention map from a single timestep of sampling instead of using the average of self-attention map across timesteps. Compared to both (1) and (2), our masks have better overlap with ODISE~\cite{xu2023open} masks while remaining confined to the plausible regions of motion (like water).}
    }
    \label{fig:mask_comparison}
    \vspace{-3mm}
    
\end{figure}

\myparagraph{Twin Image Synthesis} Here, we study the role of twin image synthesis. 
We compare our full method, which predicts optical flow on the realistic image, to Ours (w/o twin), which directly predicts optical flow using the artistic image. We also use the same mask in both cases (generated using the method mentioned in \refsec{mask}). 
\reffig{ablation}b shows that our predicted flow (using a realistic image) is significantly smoother and consistent. In our user study, $64.59 \pm 9.86 \%$ of users prefer our full method to Ours (w/o twin), which suggests the crucial role of twin image synthesis. 
%

\myparagraph{\camera{Mask Generation}} \camera{We evaluate the design choices in our mask generation method against the following alternatives : (1) using K-Means~\cite{lloyd1982least} clustering instead of spectral clustering~\cite{ng2001spectral}, (2) using self-attention map from a single timestep of sampling instead of using the average of self-attention map across timesteps. \reffig{mask_comparison} highlights that masks generated by our method have better overlap with ODISE~\cite{xu2023open} masks and are restricted to regions of motion (like water) compared to both (1) and (2). For quantitative evaluation, we manually annotated the masks for 10 randomly selected artistic images and generated a mask using all three methods. Our method achieves a better average IoU of 0.84 than (1) 0.81 and (2) 0.79.}

\subsection{Extentions}

\begin{figure}
    \centering	
    \animategraphics[autoplay,loop,width=0.475\textwidth, trim=5 0 10 0, clip]{30}
    {images/gif_controllable2/}{0}{59}
    \caption{\camera{\textbf{Text-Guided Direction Control.} The video results of our text-guided direction control using two distinct examples of different input text direction of motion for the same scene. To view this figure as a video, we recommend using Adobe Acrobat. Please see our supplementary website for more video results.}}
    \label{fig:control}
\end{figure}

\myparagraph{Text-Guided Direction Control} We can also control the motion based on text directions.
Users can indicate the direction with the template phrase, ``\textit{in ... direction}''  following each object. For example, in \reffig{control}, a user can add ``\textit{in left to right, downwards direction}'' after ``\textit{river}''. We divide the total $360^\circ$ possible directions into 12 quadrants and associate each with a direction phrase. 
We then randomly sample an angle, $\bs\theta$, from the corresponding quadrant. This angle, combined with the binary mask $\artmask$, generates a flow hint map where the $\bs y$ component is $-\sin{\bs\theta}$, and $\bs x$ component is $\cos{\bs\theta}$. Inspired by 
\cite{mahapatra2022controllable}, we extend our model to accept these hints, in addition to the realistic image $\art$, mask $\artmask$, and text prompt $\bs{c}$, to predict the flow. \reffig{control} shows two distinct examples where we can synthesize the same scene with different motion directions. 
We attempted end-to-end training to directly condition optical flow on full sentences containing direction phrases. However, this approach was unsuccessful, likely due to the limited scale of the dataset.



\myparagraph{Real Painting Cinemagraphs} Our method can also create cinemagraphs for real paintings drawn by artists, like Ivan Aivazovsky's \textit{`The Ninth Wave'} painting. We convert the real image into a natural version by Diffusion-based image editing method, Plug-and-Play~\cite{tumanyan2022plug}. Specifically, we perform DDIM inversion~\cite{song2021ddim} for 1000 steps to achieve high-quality reconstruction followed by sampling and saving the intermediate features of the Stable Diffusion model. 
The rest of the steps are similar to our core algorithm. 
\reffig{painting} shows two cinemagraphs created from historic paintings.








\section{Limitations and Discussion} \label{sec:conclusion}
\begin{figure}[t!]
    \centering
    \includegraphics[width=0.9\linewidth]{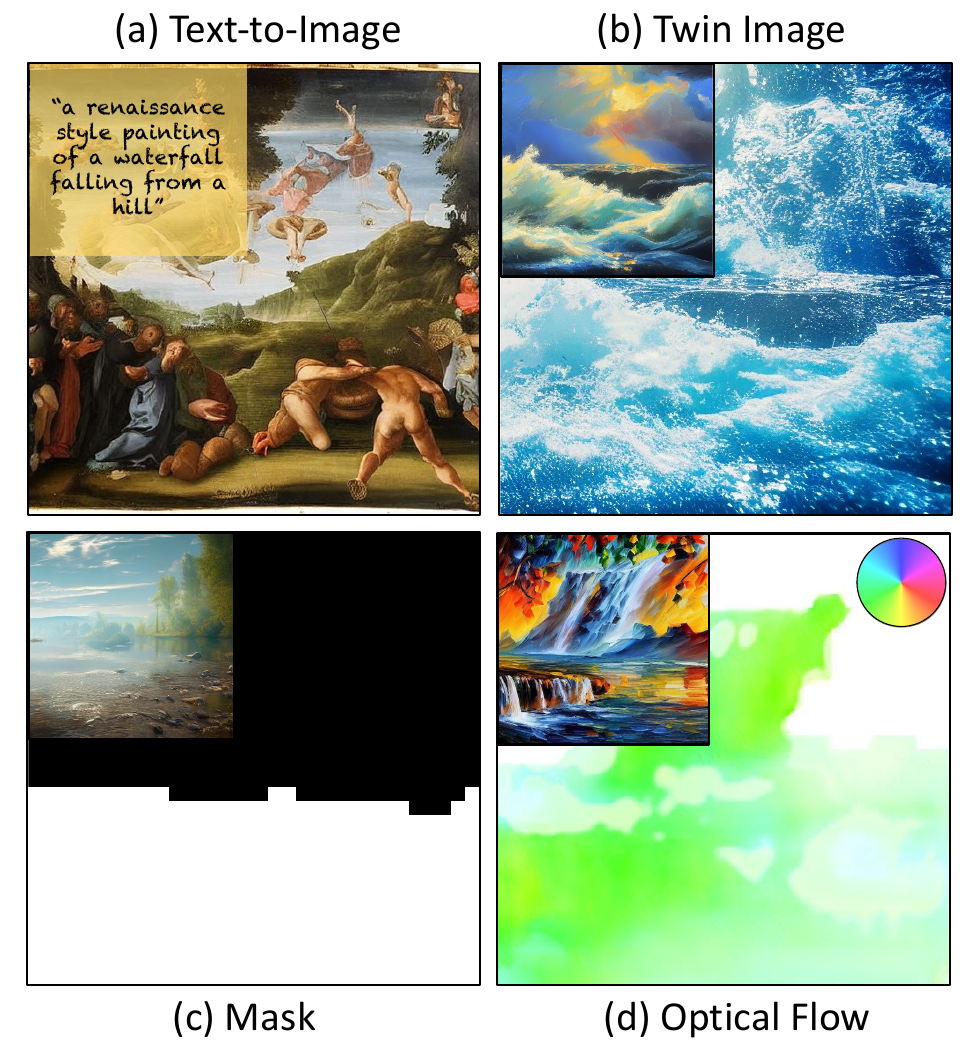}
    \caption{
    \camera{{\bf Limitations.} Our method may fail due to several reasons, including (a) incorrect text-to-image generation result, (b) inconsistency between artistic and natural images, (c) imperfect segmentation for challenging natural images, and (d) scenes with complex fluid dynamics. }
    }
    \label{fig:limitations}
\end{figure}

In summary, we have introduced the problem of creating cinemagraphs from a text description. We have presented the first fully automatic method that works well for both natural and artistic scenes. \camera{Our method not only can generate cinemagraphs for text descriptions of real-life structures with different artistic styles but also for imaginative scenes altogether for different types of fluid elements like clouds, water and smoke}. Our key concept, twin image synthesis, holds potential for other image synthesis tasks. By leveraging the concept of twin images, we separate the visual recognition process, which can be performed in an \emph{easier} domain with abundant datasets and pre-trained models, from the more challenging visual synthesis process.

\myparagraph{Limitations.}  Our method bears several limitations. 
First, the artistic and realistic images generated by Stable Diffusion may not always correspond to the input text prompt or contain very little or no dynamic regions. 
As shown in \reffig{limitations}a, the generated image does not contain the ``waterfall'' region mentioned in the text prompt.

Second, our method occasionally alters the artistic image's structure even though it shares the same self-attention maps with the realistic image. For example, as seen in \reffig{limitations}, the sky region in the artistic image has inaccurately transformed into a wave 
Using more advanced content-preserving image editing methods might be an option to alleviate the issue. 

Third, our pre-trained segmentation model, like ODISE, can struggle with complex natural images. In \reffig{limitations}c, ODISE has difficulty separating small, isolated rocks from the river.  Similarly, the optical flow network may introduce errors for natural images with unusual compositions and layouts, such as a waterfall appearing in the sky. 

Fourth, for significant changes in the flow direction, like repeated zig-zag movement of water (\reffig{limitations}d), the optical flow model may fail to predict a plausible flow. Controllable image animation models~\cite{fan2022simulating} could potentially mitigate the issue, with the user annotation cost regarding flow directions and object masks.

\camera{Finally, while our method is effective for scenes with repetitive textures, such as water, the method tends to generate monotonic movements due to symmetric splatting and the constant optical flow assumption. Additionally, our results sometimes contain artifacts of greyish regions due to the incapability of the video generator to fill holes created during symmetric splitting. 
}




\section{Acknowledgement}
We thank Gaurav Parmar and Or Patashnik for proof-
reading our manuscript. We are also grateful to Nupur Ku-
mari, Gaurav Parmar, Or Patashnik, Songwei Ge, Sheng-
Yu Wang, Chonghyuk (Andrew) Song, Daohan (Fred) Lu,
Richard Zhang, and Phillip Isola for fruitful discussions.
This work is partly supported by Snap Inc. and was partly
done while Aniruddha was an intern at Snap Inc.


\typeout{}
{\small
\bibliographystyle{ieee_fullname}
\bibliography{main}
}



\newpage

\clearpage
\appendix
\noindent{\Large\bf Appendix}
\vspace{5pt}


Please refer to our project \website for more video results and comparisons.
In \refsec{s2}, we begin by describing additional implementation details in our method. \refsec{s3} describes why data augmentation to train $\gf$ on artistic style data is not an effective approach. In \refsec{s4} we describe the details of our user study experiment. \camera{In \refsec{s8} we describe why we need to use different masks at train and test times for optical flow prediction. \refsec{s6} mentions some of the limitations in terms of capability with our current text-guided direction control approach. \refsec{s7} describes the additional limitations of the quality of cinemagraphs obtained from the baseline methods.}  Finally, in \refsec{s5} we describe why we do not use cross-attention maps instead of self-attention maps for generating the binary mask. 


\section{Additional Implementation Details} \label{sec:s2}
\vspace{2pt}
\myparagraph{Mask Generation} For generating the average self-attention map $\attenavg$, we only use the self-attention maps after $25$-th step in the denoising process, considering we perform DDIM sampling for $50$ steps. This is because the self-attention maps generated during the early phase of the denoising process are very noisy. Using them reduces the actual important semantic information for clustering. In our implementation, we use maps of resolution $32\times 32$ following Patashnik et al. \cite{patashnik2023localizing}. This is because the lower resolution maps would generate very coarse clusterings and hence very coarse binary masks, which when upsampled to image resolution of $512\times 512$, will be extremely coarse. Ideally, we would have preferred to use self-attention maps of the highest resolution possible, i.e., $64\times64$, but we observe that they do not contain very useful information for clustering. For clustering, we use 10 clusters which work well in most of the cases. For text prompts that generate very fine animatable regions (like thin waterfalls), a user might need to use a higher number of clusters. For selecting which regions to take from the cluster, we use an overlap of $\geq 70$ \% pixel-level overlap for most scenes (like river, ocean, sea, and clouds), and use a $\geq 90$ \% pixel-level overlap for waterfall scenes, which contain very fine structures.
\myparagraph{Training} During the independent first stage training of both the flow prediction and frame generation networks, we use a learning rate of $2 \times 10^{-3}$ and use the TTUR method to update the learning rate~\cite{heusel2017gans}.  For the end-to-end training, we reduce the learning rate to $1 \times 10^{-3}$ for training for an additional 50 epochs. 
We perform data augmentation, by randomly cropping the frames and optical flow to $512 \times 512$ resolution and applying random horizontal flip. We normalize the ground truth flows to $[-1, 1]$, by diving all the optical flows in the training data by a constant factor (64 in our case). We saw that normalizing the flows and using $\tanh$ in $\gf$ produces much more accurate optical flows compared to unbounded regression. We train all the networks with a batch size of 16 (both during independent and end-to-end training). While end-to-end training, we freeze the flow prediction network, $\gf$, and only fine-tune the frame generation network $\gv$.
\begin{figure}[t!]
    \centering
    \includegraphics[width=0.9\linewidth]{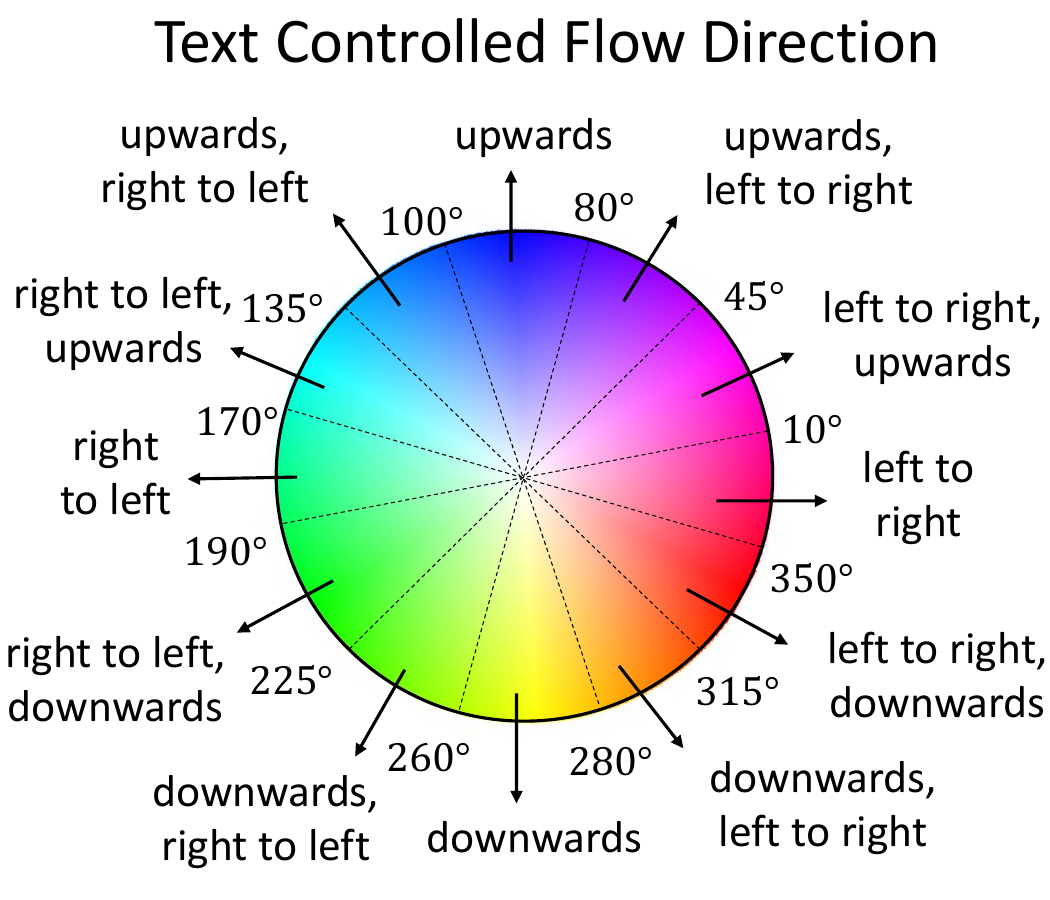}
    \caption{
    \camera{{\bf Text-guided direction control.} The wheel shows the division of all possible directions of motion in 12 quadrants and the text direction templates associated with each of the quadrants}.
    }
    \label{fig:controlwheel}
\end{figure}

\myparagraph{Inference} For inference, we clip the predicted optical flow values between $[-1,1]$ for both artistic and real domain experiments, even for baseline methods. For the artistic domain, we also scale the predicted optical flow values by a factor of $0.5$. This ensures that there do not exist very high values of optical flow which might result in large holes after symmetric splatting, which even the $\gv$ would not be able to impaint. Also, this ensures slow-moving and aesthetic structures in the generated cinemagraphs.
\begin{figure*}[t!]
    \centering
    \includegraphics[width=\linewidth]{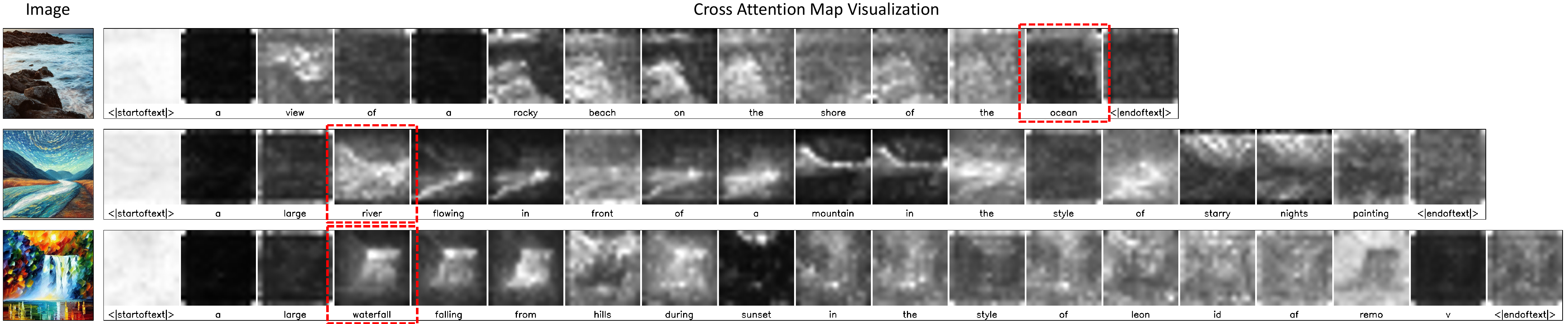}
    \caption{
    {\bf \camera{Cross Attention Maps Limitations.}} \camera{This figure visualizes the synthesized image (left) using Stable Diffusion and visualization of their corresponding average cross attention maps for each token in the input prompt (right). It can be seen that even though cross attention maps can generate a mask for the `waterfall' region properly corresponding to the `waterfall' token (bottom), for `ocean' (top) and `river' (middle), they also highlight neighboring regions of `rocky beach' and `mountain' respectively. This reduces their effectiveness in mask generation.}
    }
    \label{fig:cross_attn_maps}
\end{figure*}

\section{Why not train with Artistic Data Augmentation?}
\label{sec:s3} 
In our method, we have to generate a twin realistic image $\nat$ that shares the same semantic structure as the artistic image $\art$, which is essential for precise flow prediction, as the optical network can also work on real videos. Instead of going via this route, can we just create an augmented dataset of artistic style images and optical flow by editing the real images, using various diffusion-based semantically aligned editing methods 
into various artistic styles? These augmented artistic style images will essentially have the same ground-truth flow as the real images from which they were created as these editing methods, for most cases, preserve semantic layout in the edit. We, however, have tried this approach but failed to generate accurate flows at inference for artistic images. We believe there are mainly two reasons, (1) the number of artistic styles can be extremely large, and generating augmentations and further training with all possible augmentations like extremely resource and time-intensive, at inference, the text prompt might contain an artistic style very different from training augmentations, (2) The augmentations are limited by the structures of the real domain images. The text prompts at inference can not only contain artistic styles but also outwardly structure, in which case this method also fails.
\section{User Study Details} \label{sec:s4} For both real and artistic domain experiments we perform a user study to evaluate the quality of the generated cinemagraphs. We believe that to assess the quality of generative videos (and also cinemagraphs), human evaluation is the best available metric. We perform human evaluations on Amazon Mechanical Turk. We performed a paired test, where the users were asked to compare two videos, one generated by our method and the other by one of the baselines or our ablation methods, based on either a single image (for the real domain) or a text prompt (for the artistic domain). Each paired comparison was annotated by $5$ annotators. Therefore, we collect a total of $810$ responses for the real domain (considering 162 test images) and $500$ responses for the artistic domain (considering 100 unique text-seed pairs). Only annotators who had more than $10000$ accepted hits and had an accuracy of more than $95$\% could participate in the task. Each annotator received an amount of \$0.03 for one annotation (or hit). Based on the obtained hits, for each of our v/s baseline (or our v/s  ablation) tests, we bootstrap the samples 1000 times and report the mean and standard error in Table \ref{tab:user_study} in the main paper. 
\section{Mask for training}
\label{sec:s8} 
\camera{As mentioned in Section 3.2 in the main paper, we need a mask in addition to the single image to train the flow prediction network $\gf$. Since we train $\gf$ only on real domain data, one option could be to use a mask obtained by a pretrained image segmentation model, like ODISE \cite{xu2023open}, similar to how we generate mask a test time for real domina. Instead, at training time, we use masks generated by thresholding the ground-truth average optical flow (like \cite{mahapatra2022controllable, fan2022simulating}). We find that training with masks obtained by thresholding the ground-truth average optical flow, makes $\gf$ learn to predict non-zero optical flow in all regions inside the mask. In contrast, we observe that for the real domain training dataset, the masks generated by ODISE often include more regions that can have zero ground-truth average optical flow. This might be due to the inability of the pretrained optical flow model used to estimate optical flow in ground-truth videos in regions of (relatively) less movement. Thus using ODISE-generated masks during training will not ensure that $\gf$ learns to predict non-zero optical flow in all regions inside the mask. In our experimentation, we saw this leads to inferior results compared to training with masks obtained by thresholding ground-truth average optical flow.}
\section{Limitations of Text-Guided Direction Control } 
\label{sec:s6} 
\camera{In Section 4.5 of the main paper, we show that our method of generating cinemagraphs from text prompts can be extended to controlling the direction of motion in generated cinemagraph using text directions in the input prompt. However, to this end, our method only allows the user to specify a global direction. This works well for most scenarios where there is a single body that moves relatively in the same direction(like ocean or clouds or one waterfall or river). But, our method of text-guided direction control won't work in scenarios where the user wants to (1) generate two or more objects (like rivers) moving in different directions, and, (2) assign different local directions of motion within the same object. We leave this as future work to investigate a way to achieve more fine-grained and end-to-end text-guided direction control.}

\section{Limitations of Baselines} \label{sec:s7} 
\camera{The text-to-video baselines, CogVideo \cite{hong2022cogvideo}, Text2Video-Zero \cite{khachatryan2023text2video} and VideoCrafter~\cite{videocrater} natively generates videos of $32$, $8$, and $16$ frames at resolutions of $480\times 480$, $512 \times 512$, and $256 \times 256$,  respectively. Although they can generate videos of $120$ frames (the same number of frames as our method), we found that increasing the number of generated frames greatly reduces the temporal consistency of the output videos from these methods. Thus, we compare the videos generated by these baseline methods against our method in their respective native resolution and frame number.
}

\section{Cross-Attention Maps v/s Self-Attention maps for Mask Generation} \label{sec:s5} We use self-attention maps instead of cross-attention maps for generating the binary mask $\bs{M}$. Even though Prompt-to-Prompt~\cite{hertz2022prompt} shows that cross-attention map from the intermediate layers in Stable Diffusion for a particular token generates a mask that is representative of what that token corresponds to in the output image. \camera{In \reffig{cross_attn_maps} we see that this is not always the case. For scenes with larger or spread-out structures, like oceans and rivers, the cross-attention map also tends to highlight neighboring regions.}

\end{document}